\documentclass[runningheads]{llncs}
\usepackage{graphicx}
\usepackage{multirow}
\usepackage{booktabs}
\usepackage{amsfonts} 
\usepackage{gensymb}
\usepackage{float}
\usepackage{capt-of}
\usepackage{centernot}
\usepackage{mathtools}
\usepackage{stmaryrd}
\usepackage{pifont}
\usepackage{hyperref}
\usepackage{subcaption}

\begin{document}
\title{SR4ZCT: Self-supervised Through-plane Resolution Enhancement for CT Images with Arbitrary Resolution and Overlap}
\titlerunning{SR4ZCT}
%
\author{Jiayang Shi\inst{1} \and
Dani\"{e}l M. Pelt\inst{1} \and
K. Joost Batenburg\inst{1}}

 \authorrunning{J. Shi et al.}
%
\institute{Leiden University, Leiden, the Netherlands\inst{1}\\ \email{\{j.shi, d.m.pelt, k.j.batenburg\}@liacs.leidenuniv.nl}\\}

\maketitle              
\begin{abstract}
Computed tomography (CT) is a widely used non-invasive medical imaging technique for disease diagnosis. The diagnostic accuracy is often affected by image resolution, which can be insufficient in practice. For medical CT images, the through-plane resolution is often worse than the in-plane resolution and there can be overlap between slices, causing difficulties in diagnoses. Self-supervised methods for through-plane resolution enhancement, which train on in-plane images and infer on through-plane images, have shown promise for both CT and MRI imaging. However, existing self-supervised methods either neglect overlap or can only handle specific cases with fixed combinations of resolution and overlap. To address these limitations, we propose a self-supervised method called SR4ZCT. It employs the same off-axis training approach while being capable of handling arbitrary combinations of resolution and overlap. Our method explicitly models the relationship between resolutions and voxel spacings of different planes to accurately simulate training images that match the original through-plane images. We highlight the significance of accurate modeling in self-supervised off-axis training and demonstrate the effectiveness of SR4ZCT using a real-world dataset.

\end{abstract}
\section{Introduction}

CT is a valuable tool in disease diagnosis due to its ability to perform non-invasive examinations \cite{hansen2021computed}. The diagnostic accuracy of CT imaging is dependent on the resolution of the images, which can sometimes be insufficient in practice \cite{iwano2004solitary}. For medical CT images, the through-plane resolution is often inferior to the in-plane axial resolution \cite{tsukagoshi2007improvement,angelopoulos2012comparison}, resulting in anisotropic reconstructed voxels that can create difficulties in identifying lesions and consequently lead to inaccurate diagnoses \cite{coward2014multi,he2016effects,angelopoulos2012comparison}. Furthermore, medical CT scans are commonly performed in a helical trajectory, which can cause overlapping slices along the through-plane axis \cite{kasales1995reconstructed,brink1995technical}. The overlap caused by certain combinations of through-plane resolution and spacing introduces extra blurriness in sagittal and coronal images and potentially leads to inaccuracies in image interpretation \cite{honda2007computer,ravenel2008pulmonary,gavrielides2013benefit}.

To improve the through-plane resolution of CT images, various deep learning-based methods have been proposed. Supervised methods that use pairs of low- (LR) and high-resolution (HR) volumes have demonstrated promising results \cite{peng2020saint,liu2020multi,yu2022rplhr}. However, HR reference volumes may not always be available. As an alternative, self-supervised methods have been introduced, training on in-plane images and applying the learned model to through-plane images \cite{xie2021high,zhao2020smore}. These methods simulate LR training images from HR in-plane images that match the through-plane images. However, they have limitations in handling CT images with overlap or complicated resolution/overlap combinations. For example, \cite{xie2021high} can only handle CT images without overlap and integer resolution ratios,
while the use of convolutional operations in \cite{zhao2020smore} restricts its usage to certain specific combinations of overlap and resolution.

In this work, we propose a self-supervised method for enhancing the through-plane resolution of CT images by explicitly modeling the relation between the resolutions and spacings of different planes. Our method builds upon the same idea of off-axis training introduced in \cite{xie2021high,zhao2020smore}, but with the ability to handle CT images with arbitrary resolution/overlap. We use accurate interpolation techniques to simulate LR images from HR images, considering the specific resolution and overlap parameters. Our contributions are: (1) proposing a self-supervised method that can enhance resolution for CT images with arbitrary resolution and overlap, (2) demonstrating the importance of accurate modeling in off-axis training, and (3) applying our method to real-world datasets with complicated resolution/overlap combinations, showcasing its practical applicability.


\section{Method}
\begin{figure}[t]
\centering
\includegraphics[width=1\textwidth]{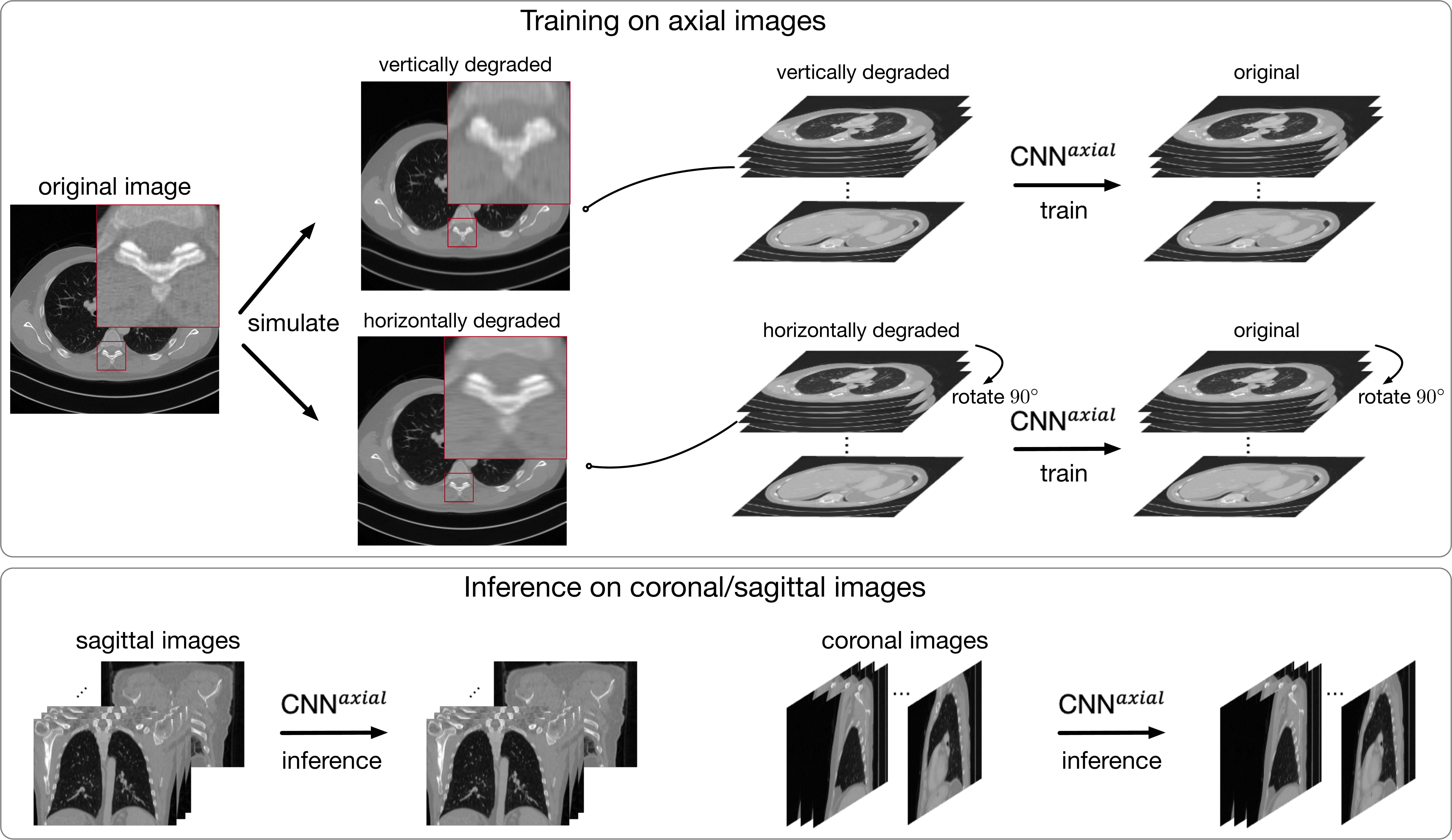}
\caption{Overview of our SR4ZCT method.}
\label{method}
\includegraphics[width=1\textwidth]{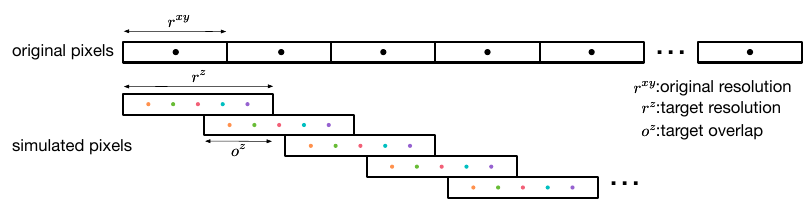}
\caption{$\mathcal{F}^{ver}_{\downarrow}$ to simulate abitrary resolution and overlap. The function linearly interpolates at various points (indicated by different colors) inside the pixels, these interpolated points are then averaged to be the values of the current pixels.}
\label{interpolation}
\end{figure}

\textbf{Problem Statement and Method.} We present in this part a generalized problem statement for enhancing the resolution of CT images and describe our proposed method SR4ZCT as illustrated in Figure \ref{method}. We denote the reconstructed CT volume by $\mathbf{I}(x,y,z) \in \mathbb{R}^{X \times Y \times Z}$, where $X, Y, Z$ represent the pixel numbers along $x, y, z$ axes, respectively. CT images can be viewed along three different orientations: axial, coronal, and sagittal. We denote the axial image, the coronal image, and the sagittal image at the position $z'$, $y'$, and $x'$ along the $z$ axis, the $y$ axis, and the $x$ axis as $\mathbf{a}_{z'}=\mathbf{I}(:,:,z')$, $\mathbf{c}_{y'}=\mathbf{I}(:,y',:)$, and $\mathbf{s}_{x'}=\mathbf{I}(x',:,:)$, respectively. The voxel size of the CT volume is defined as $r^{x}\times r^{y}\times r^{z}$, where often $r^{x}=r^{y}=r^{xy}<r^{z}$ for medical CT images. The spacing distance $d$ between the centers of neighboring voxels along the $z$-axis can be smaller than $r^{z}$, resulting in the overlap $o^{z}=r^{z}-d$ in the through-plane direction. Given the target voxel size $r^{x}\times r^{y}\times r^{z}_{tar}$ and overlap $o^{z}_{tar}$, the goal of resolution enhancement for CT through-plane images is to reduce their voxel size and overlap in $z$ direction, hereby increasing the through-plane resolution. 


SR4ZCT is based on generating a training dataset that consists of HR axial images as targets and their corresponding virtual through-plane-like axial images as inputs. To achieve this, we define the downscaling function $\mathcal{F}^{ver}_{\downarrow}$ as shown in Figure \ref{interpolation}, which downscales an axial image $\mathbf{a}_i$ with pixel sizes $r^{xy} \times r^{xy}$ by employing linear interpolation in the vertical direction. The output is a virtual through-plane-like image with pixel sizes $r^z \times r^{xy}$ and overlaps $o^z$ between image rows. We similarly define the upscaling function $\mathcal{F}^{ver}_{\uparrow}$ that upscales the virtual through-plane-like image with pixel sizes $r^z \times r^{xy}$ and overlap $o^z$ using linear interpolation to produce a degraded axial image with pixel sizes $r^{xy} \times r^{xy}$ and zero overlap. In addition, we define scaling functions $\mathcal{F}^{hor}_{\downarrow}$ and $\mathcal{F}^{hor}_{\uparrow}$ that apply the scaling in the horizontal direction instead of the vertical direction. 

Training data is produced by first creating two network input images for each axial image $\mathbf{a}_i$: $\mathbf{a}_i^{ver} = \mathcal{F}^{ver}_{\uparrow}( \mathcal{F}^{ver}_{\downarrow}(\mathbf{a}_i))$ and $\mathbf{a}_i^{hor} = \mathcal{F}^{hor}_{\uparrow}( \mathcal{F}^{hor}_{\downarrow}(\mathbf{a}_i))$. Next, the simulated images are fed into the neural network $f_{\theta}$ to learn the mapping from degraded images $\mathbf{a}_i^{ver}$ and $\mathbf{a}_i^{hor}$ to their corresponding original axial image $\mathbf{a}_i$. As the resolution enhancement is along the $z$ axis, which is always in the vertical direction for coronal and sagittal images, the horizontally degraded images and their corresponding axial images are rotated $90\degree$ using rotation function $\mathcal{R}$. The training is performed on all $Z$ axial images and their corresponding degraded images using the loss function $L$, resulting in $2\times Z$ training pairs. The weights $\theta$ of the neural network $f_{\theta}$ are determined by minimizing the total loss as 
\begin{equation}
\label{eqn:deeplearningobjective}
\theta^{*} = \min_{\theta} \sum_{i=1}^{Z} L(f_{\theta}(\mathbf{a}_i^{ver}),\mathbf{a}_i) + L(f_{\theta}(\mathcal{R}(\mathbf{a}_i^{hor})),\mathcal{R}(\mathbf{a}_i)).
\end{equation}

After training, the network $f_{\theta}$ represents the learned mapping from virtual through-plane-like axial images with resolution $r^{z}\times r^{xy}$ and overlap $o^{z}$ between image rows to HR axial image with $r^{xy}\times r^{xy}$ resolution. With the assumption that CT images often share similar features from different orientations, we can extend the learned mapping from axial images to coronal and sagittal images. To enhance the resolution, we first use $\mathcal{F}^{ver}_{\uparrow}$ to upscale coronal and sagittal images with voxel sizes $r^{z}\times r^{xy}$ and overlap $o^{z}$ between image rows to images with resolution $r^{xy}\times r^{xy}$. Subsequently, we apply the trained neural network $f_{\theta}$ directly to the upscaled coronal and sagittal images. The outputs of the neural network correspond to the improved images with enhanced resolution.

In general, any image-to-image neural network could be used as $f_{\theta}$. In this work, we use 2D MS-D networks \cite{pelt2018mixed} with 100 layers, trained for 200 epochs using L2 loss and ADAM optimizer \cite{kingma2014adam}. The code is avaiable on GitHub. \footnote{\href{https://github.com/jiayangshi/SR4ZCT}{https://github.com/jiayangshi/SR4ZCT}}

\section{Experiments and Results}

\textbf{Comparison with Supervised Learning.} We conducted an experiment to evaluate the effectiveness of SR4ZCT in improving the resolution of CT images, comparing it to the supervised learning method. We selected four volumes (nr. 06, 14, 22, 23) from the \texttt{Task06 Lung} dataset of the Medical Segmentation Decathlon \cite{simpson2019large}. These volumes had the same slice thickness ($0.625 mm$) but varying axial resolutions ranging from $0.74 mm$ to $0.97 mm$, resulting in a total of 2170 axial images of size $512 \times 512$ pixels. We downscaled the axial images to a resolution of $1 mm$, creating volumes with voxel sizes of $1 mm \times 1 mm \times 0.65 mm$. The through-plane resolution of the simulated volumes ranged from $2.5 mm$ with $1.25 mm$ overlap to $6.25 mm$ with $3.125 mm$ overlap. We trained the supervised learning method using three LR and HR volume pairs (numbers 06, 14, 22), and evaluated it on the remaining LR CT volume (nr. 23). We utilized two distinct 2D MS-D networks to train the coronal and sagittal images separately in a supervised manner, thereby achieving the best possible results through supervised learning. We applied SR4ZCT to the LR CT volume (nr. 23) and compared the results with those obtained using the supervised learning method.

\begin{table}[t]
\setlength{\tabcolsep}{3pt}
\centering
\begin{tabular}{ccccccccc} \toprule[0.1pt]
                \multirow{2}{*}{resolution} & \multirow{2}{*}{overlap}  & \multirow{2}{*}{view} & \multicolumn{2}{c}{original} & \multicolumn{2}{c}{supervised}  &  \multicolumn{2}{c}{ours}\\
                & & & PSNR & SSIM & PSNR & SSIM & PSNR & SSIM\\\midrule[0.05pt]
\multirow{2}{*}{$2.5 mm$} & \multirow{2}{*}{$1.25 mm$} & cor & 37.71 & 0.972 & \textbf{51.94} & \textbf{0.998} & 46.48 & 0.995 \\
                  & & sag & 39.12 & 0.977 & \textbf{52.50} & \textbf{0.998} & 47.68 & 0.995 \\ \midrule[0.05pt]
\multirow{2}{*}{$3.75 mm$} & \multirow{2}{*}{$1.875 mm$} & cor & 33.65 & 0.934 & \textbf{42.96} & \textbf{0.987} & 42.36 & 0.985 \\
                  & & sag & 35.08 & 0.945 & \textbf{43.73} & \textbf{0.989} & 43.39 & 0.987 \\ \midrule[0.05pt]
\multirow{2}{*}{$5 mm$} & \multirow{2}{*}{$2.5 mm$} & cor & 31.43 & 0.895 & 37.70 & 0.965 & \textbf{38.46} & \textbf{0.967} \\
                  & & sag & 32.87 & 0.913 & 39.36 & \textbf{0.973} & \textbf{39.55} & 0.973 \\ \midrule[0.05pt]
\multirow{2}{*}{$6.25 mm$} & \multirow{2}{*}{$3.125 mm$} & cor & 29.91 & 0.9859 & 34.74 & 0.938 & \textbf{35.88} & \textbf{0.946} \\
                  & & sag & 31.32 & 0.882 & 36.02 & 0.949 & \textbf{36.96} & \textbf{0.955} \\ \midrule[0.05pt]
\end{tabular}
\caption{Comparison of SR4ZCT and supervised learning on simulated CT volumes. The used metrics are PSNR and SSIM calculated from coronal and sagittal images in the central area, after cropping the empty regions.}
\label{ourandsupervised}

\centering
\includegraphics[width=1\textwidth]{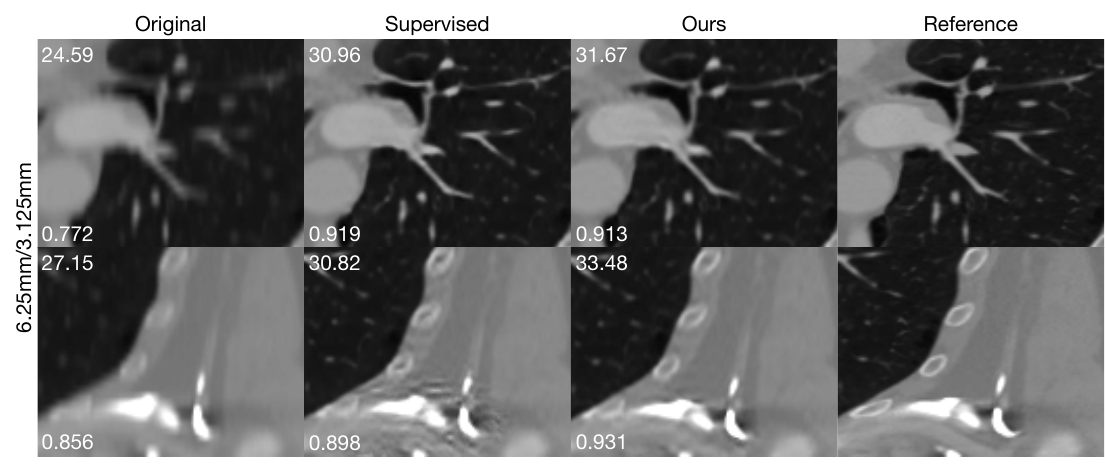}
\captionof{figure}{The comparison of supervised learning and SR4ZCT on $6.25mm$ resolution and $3.125mm$ overlap dataset. Patches are shown for visualization. The PSNR and SSIM of each patch are shown in the top- and bottom-left corners.}
\label{supervised}
\end{table}

Table \ref{ourandsupervised} shows that SR4ZCT successfully enhanced the resolution of all four simulated CT volumes, as indicated by the improvements in both PSNR and SSIM. While SR4ZCT showed slightly lower PSNR and SSIM values than the supervised learning method for the $2.5 mm$/$1.25 mm$ case, its performance approached and surpassed that of the supervised learning method as the resolution decreased. Specifically, for the $3.75 mm$/$1.875 mm$ case, SR4ZCT achieved similar results to supervised learning, whereas in the $5 mm$/$2.5 mm$ and $6.25 mm$/$3.125 mm$ cases, SR4ZCT outperformed supervised learning. Figure \ref{supervised} presents an example of the $6.25 mm$/$3.125 mm$ case, where both the supervised learning method and SR4ZCT performed similarly in the lung area as SR4ZCT, but supervised learning created artifacts on the edge area, highlighting one of its disadvantages as it depends on the quality and quantity of training data. If the training data slightly differs from the testing data or is insufficient in amount, it may perform suboptimally. In contrast, SR4ZCT only makes use of the same data for training and testing, reducing the potential gap between training and testing data and yielding more robust results. Overall, our results demonstrate the effectiveness of SR4ZCT in enhancing the resolution of through-plane images, with similar or even superior performance compared with supervised learning.

\textbf{Comparison with SMORE.} SMORE \cite{zhao2020smore} is a method designed for MRI images and relies on the fact that MRI acquisition is akin to sampling in the Fourier domain. It simulates LR images from HR images by removing the high-frequency components. However, SMORE is not directly applicable to CT imaging. Nonetheless, we implemented a version of SMORE based on its core idea of simulating LR images by convolving HR images. While convolution can simulate certain resolution/overlap combinations by adjusting the filter kernel and stride, it fails to accurately simulate specific resolution/overlap combinations, when the target resolution/overlap is not an integer multiple of the HR resolution/overlap.

To compare our method with SMORE, we used the same volumes of the Medical Segmentation Decathlon dataset \cite{simpson2019large} and downscaled them in the vertical direction to simulate low through-plane resolution with overlaps of $2 mm$, $2.5 mm$, and $3 mm$ for a through-plane resolution of $5 mm$. The results in Table \ref{comparewithsmore} demonstrate that SMORE performs better when the simulated LR images using convolution match the through-plane resolution/overlap. For example, applying 1D convolution with a filter width of 5 and a stride of 3 on the $1mm\times1mm$ HR axial images is equivalent to simulating a resolution of $5 mm$ and an overlap of $2 mm$. It achieved higher PSNR values on the $5 mm/2 mm$ volumes than the one using a stride of 2. When the target overlap is $2.5 mm$, no convolution configuration accurately simulates it by adjusting the stride and resulting in lower PSNRs. The same limitation applies when the ratio of actual and target resolution is non-integer, making it impossible to find an accurate convolution filter with a decimal filter width. These limitations of convolution result in suboptimal performance. In contrast, our method simulates LR images using accurate interpolation, which can handle arbitrary resolution/overlap combinations.

\begin{table}[t]
\centering
\setlength{\tabcolsep}{5pt}
\begin{tabular}{cccc} \toprule[0.1pt]
 & $5 mm$/$2 mm$ & $5 mm$/$2.5 mm$ & $5 mm$/$3 mm$ \\ \midrule[0.05pt]
SMORE stride 2 & $37.15$ & $38.39$ & $43.55$ \\ \midrule[0.05pt]
SMORE stride 3  & $41.93$ & $41.86$ & $41.41$ \\ \midrule[0.05pt]
ours & $\textbf{42.93}$ & $\textbf{43.30}$ & $\textbf{44.75}$       \\ \bottomrule[0.1pt]
\end{tabular}
\caption{Results of SMORE and our method applied on the simulated dataset. The used metric is PSNR, the values are computed based on the average PSNR of the coronal images of all testing volumes. Output images are given in the supplementary file.}
\label{comparewithsmore}

\centering
\includegraphics[width=1\textwidth]{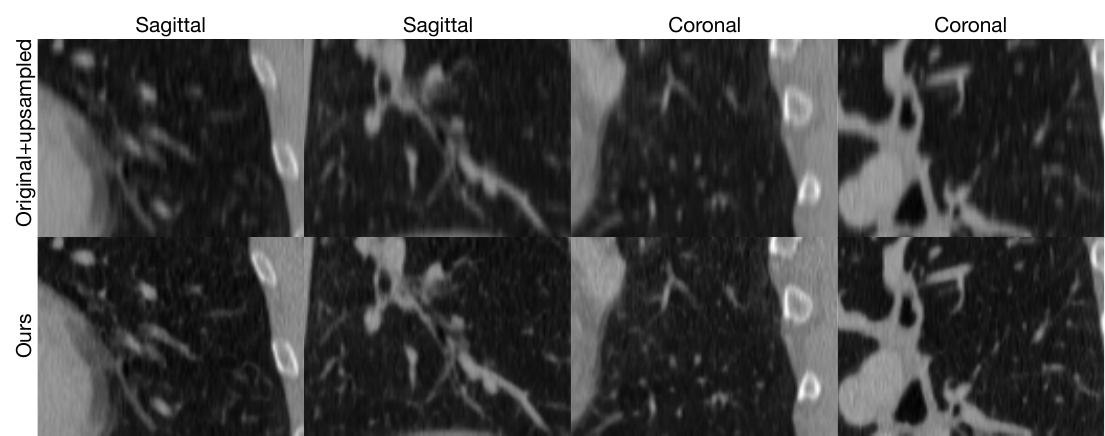}
\captionof{figure}{Results of SR4ZCT applied on \texttt{L291} from Low Dose CT Grand Challenge \cite{mccollough2017low}. We show two image patches of coronal and sagittal images. The first row contains patches of original coronal and sagittal images, and the second row shows the output of SR4ZCT. Extra results are shown in the supplementary file.}
\label{results}
\end{table}

\textbf{Real-world CT Images without Reference.} We present this experiment designed to evaluate the effectiveness of SR4ZCT on real-world CT images with anisotropic resolution and overlap. We selected a CT volume of patient \texttt{L291} from the Low Dose CT Challenge \cite{mccollough2017low}, which has an in-plane resolution of $0.74mm$, through-plane slice thickness of $3 mm$ and $1 mm$ overlap. As no reference volume without overlap was available, we present the result only for visual assessment.


Figure \ref{results} illustrates the improvement in resolution of the original coronal and sagittal images using SR4ZCT. The enhanced images exhibit sharper details compared to the original images, as observed in the visual comparison. This experiment provides evidence of SR4ZCT's effectiveness in improving CT image resolution in real-world scenarios where HR reference images are unavailable.

\textbf{Correct Modeling is Essential.} We present this experiment to demonstrate the importance of correctly modeling the simulated training images from axial images, which are used for training the neural network, to match the resolution and overlap of coronal and sagittal images. To validate this requirement, we intentionally introduced modeling errors into the training images, simulating different combinations of resolution and overlap that deviated by either $0.25mm$ or $0.5mm$ from the actual resolution of $5mm$ and $2.5mm$ overlap.

\begin{figure}[t]
\centering
\includegraphics[width=0.9\textwidth]{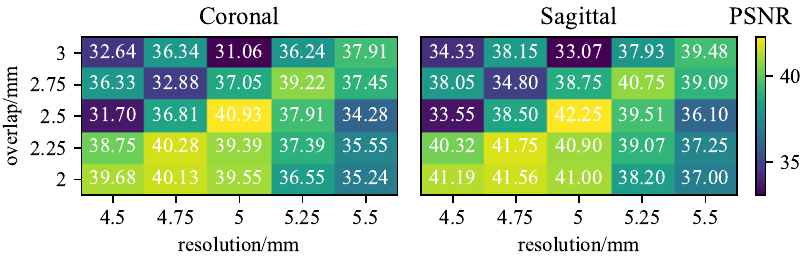}
\caption{The PSNR of cases where training images were inaccurately modeled. Each block refers to a case where the corresponding resolution/overlap are used to model the training images. The actual resolution/overlap are $5mm$/$2.5mm$.} 
\label{mismatch_plot}
\includegraphics[width=1\textwidth]{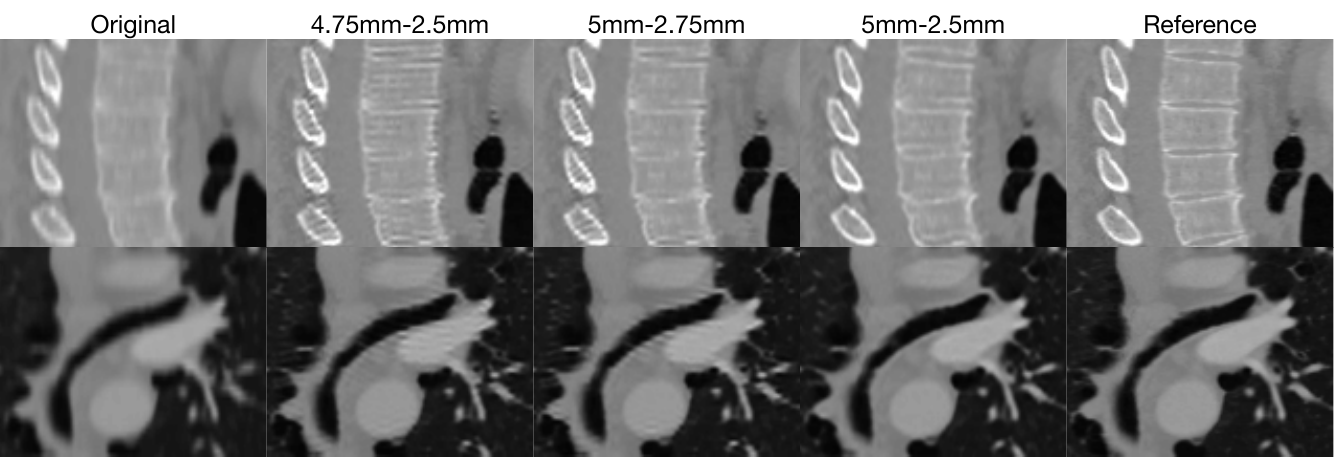}
\caption{Results of cases where training images were not correctly modeled. The actual through-plane resolution and overlap are $5mm$ and $2.5mm$. The second and third column shows cases when modeled training images deviated by $0.25mm$.} 
\label{mismatch}
\end{figure}

Fig \ref{mismatch_plot} shows SR4ZCT achieved the highest PSNR when the training images accurately reflected a resolution of $5mm$ and an overlap of $2.5mm$. Even minor deviations, such as $0.25mm$ errors, resulted in a decrease in PSNR. Interestingly, we observed that when the ratio of error in resolution and overlap was similar, the performance was better than when the error was present in only one parameter. Fig \ref{mismatch} provides examples where incorrect modeling of resolution or overlap by $0.25mm$ led to artifacts, highlighting the impact of inaccurate modeling.

\begin{figure}[t]
\centering
\includegraphics[width=\textwidth]{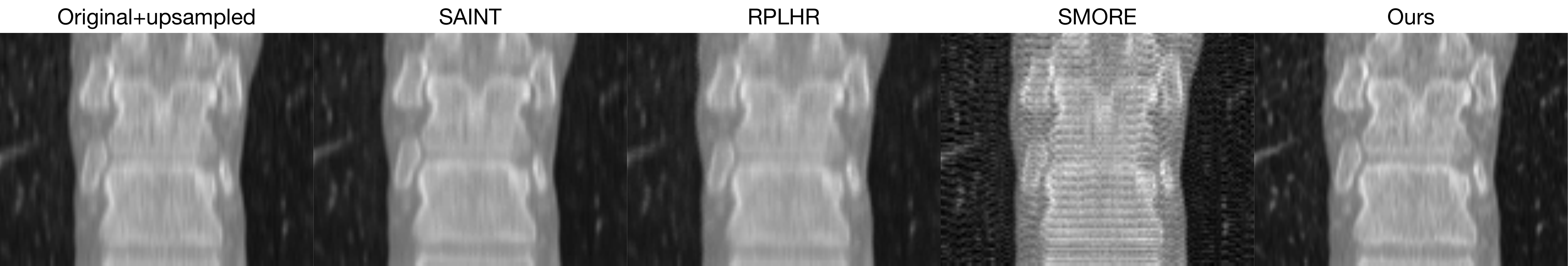}
\caption{Visual comparison of state-of-the-art CT images supervised super-resolution methods SAINT \cite{peng2020saint}, RPLHR \cite{yu2022rplhr}, self-supervised method SMORE \cite{zhao2020smore} and our method SR4ZCT applied to a real-world sagittal image.} 
\label{compare}
\end{figure}

 Fig \ref{compare} also demonstrates the importance of modeling the training images accurately. We applied supervised methods SAINT \cite{peng2020saint}, RPLHR \cite{yu2022rplhr}, and self-supervised SMORE \cite{zhao2020smore}, to Low Dose CT Grand Challenge \cite{mccollough2017low}. We used the provided pre-trained weights for SAINT and trained RPLHR as described on their dataset. SAINT and RPLHR do not consider overlap in their method design. For SMORE, we used convolution with a filter width of 4 and stride 3, which was equivalent to simulating $2.96 mm$ resolution and $0.74 mm$ overlap images, the closest possible combination to the actual $3mm$ resolution and $1mm$ overlap. The blurriness caused by the overlap in the LR image was amplified in the results of SAINT and RPLHR, while artifacts occurred in the output of SMORE. In contrast, SR4ZCT improved the resolution and reduced blurriness by accurately modeling the training images. This stresses the importance of correctly modeling training images to ensure improvement for resolution effectively.

\section{Conclusion}
In this work, we presented SR4ZCT, a self-supervised method designed to improve the through-plane resolution of CT images with arbitrary resolution and overlap. The method is based on the same assumption as \cite{xie2021high,zhao2020smore} that images of a medical CT volume from different orientations often share similar image features. By accurately simulating axial training images that match the resolution and overlap of coronal and sagittal images, SR4ZCT effectively enhances the through-plane resolution of CT images. Our experimental results demonstrated that SR4ZCT outperformed existing methods for CT images with complicated resolution and overlap combinations. It successfully enhanced the resolution of real-world CT images without the need for reference images. We also emphasized the crucial role of correctly modeling the simulated training images for such off-axis training-based self-supervised method. In the future, our method may have potential applications in a wide range of clinical settings where differences in resolution across volume orientations currently limit the 3D CT resolution. 


\section*{Acknowledgments}
This research was financed by the European Union H2020-MSCA-ITN-2020 under grant agreement no. 956172 (xCTing). 

%
%
%
\bibliographystyle{splncs04}
\bibliography{SelfSupervisedResolution.bib}
%





\end{document}